\documentclass[11pt]{article}
\usepackage{coling2020}
\usepackage{times}
\usepackage{url}
\usepackage{latexsym}
\usepackage{graphicx}
\usepackage{multirow}
\usepackage{booktabs}
\usepackage{amsmath}
\usepackage{stfloats}
\usepackage{amsfonts}
\usepackage{color,xcolor}
\usepackage{adjustbox}
\usepackage{wrapfig}
\usepackage{subfigure}

\definecolor{reds}{RGB}{184,84,80}
\definecolor{blues}{RGB}{16,115,158}

\colingfinalcopy 

\title{Bridging Text and Knowledge with Multi-Prototype Embedding for Few-Shot Relational Triple Extraction}

\author{
Haiyang Yu$^{1,2}$\thanks{\quad Equal contribution and shared co-first authorship.}, 
Ningyu Zhang$^{1,2*\dag}$,
Shumin Deng$^{1,2}$, 
Hongbin Ye$^{1,2}$, \\
\textbf{Wei Zhang}$^{3}$, 
\textbf{Huajun Chen}$^{1,2}$ \thanks{\quad Corresponding author} \\
$^{1}$ Zhejiang University \\
$^{2}$ AZFT Joint Lab for Knowledge Engine \\
$^{3}$  Alibaba Group \\
 {\tt \{yuhaiyang,zhangningyu,231sm,yehb,huajunsir\}@zju.edu.cn}\\
 {\tt lantu.zw@alibaba-inc.com} 
}

\date{}

\begin{document}

\maketitle

\begin{abstract}
Current supervised relational triple extraction approaches require huge amounts of labeled data and thus suffer from poor performance in few-shot settings. However, people can grasp new knowledge by learning a few instances. To this end, we take the first step to study the few-shot relational triple extraction, which has not been well understood. Unlike previous single-task few-shot problems, relational triple extraction is more challenging as the entities and relations have implicit correlations. In this paper, We propose a novel multi-prototype embedding network model to jointly extract the composition of relational triples, namely, entity pairs and corresponding relations. To be specific, we design a hybrid prototypical learning mechanism that bridges text and knowledge concerning both entities and relations. Thus, implicit correlations between entities and relations are injected. Additionally, we propose a prototype-aware regularization to learn more representative prototypes. Experimental results demonstrate that the proposed method can improve the performance of the few-shot triple extraction. 
\end{abstract}

\section{Introduction}

Relational Triple Extraction is an essential task in Information Extraction for Natural Language Processing (NLP) and Knowledge Graph (KG) \cite{Yu2017ImprovedNR,Huang2020KnowledgeGA}, which is aimed at detecting a pair of entities along with their relation from unstructured text. For instance, there is a sentence ‘‘\emph{Paris is known as the romantic capital of France.}'', and in this example, an ideal relational triple extraction system should extract the relational triple {\it $\langle$Paris, Capital\_of, France$\rangle$}, in which {\it Capital\_of} is the relation of {\it Paris} and {\it France}.

Current works in relational triple extraction typically employ traditional supervised learning based on feature engineering \cite{Kambhatla2004CombiningLS,Reichartz2010SemanticRE} and neural networks \cite{Zeng2014RelationCV,Bekoulis2018AdversarialTF}. The main problem with supervised learning models is that they can not perform well on unseen entity types or relation categories (e.g., train a model to extract knowledge triples from the economic text, then run this model to work on scientific articles). As a result, supervised relational triple extraction can not extend to the unseen entity or relation types. A trivial solution is to annotate more data for unseen triple types, then retraining the model with newly annotated data \cite{Zhou2019JointEA}. However, this method is usually impractical because of the extremely high cost of annotation.

\begin{figure}
  \centering 
  \includegraphics[width=0.8\textwidth]{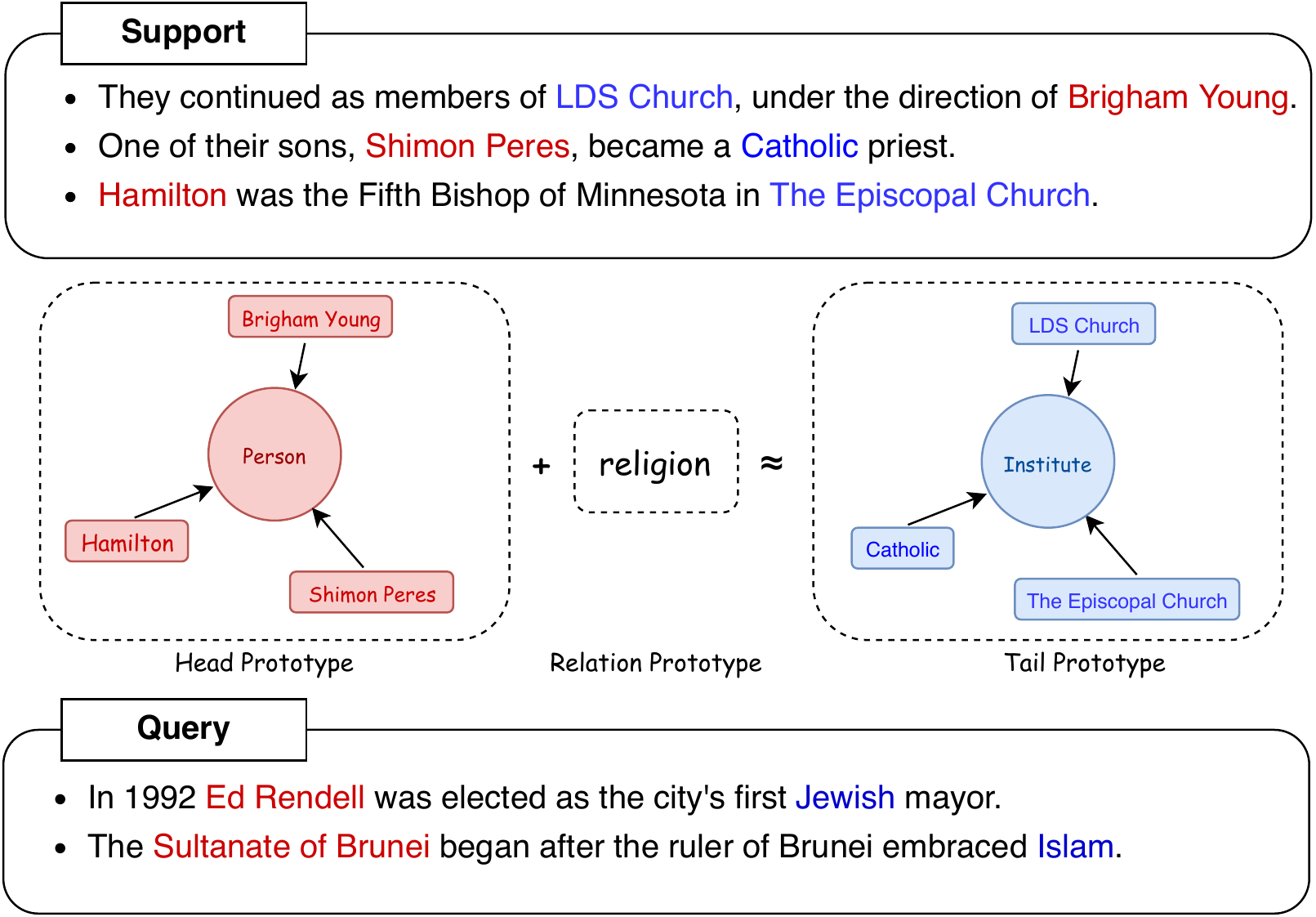} 
  \caption{Illustration of our proposed model for relational triple extraction in the few-shot setting. The texts marked in {\color{red}red} are head entities while in {\color{blue}blue} are tail entities. Head and tail entity prototypes are connected with the relation prototype.}
  \label{figure:intro}
\end{figure}

Intuitively, humans can learn about a new concept with limited supervision, e.g., one can detect and classify new entities with 3-5 examples \cite{Grishman2005NYUsEA}. This motivates the setting that we aim at for relational triple extraction: Few-Shot Learning  (FSL). In few-shot learning, a trained model rapidly learns a new concept from a few examples while keeping great generalization from observed examples \cite{Vinyals2016MatchingNF,deng2020low}. Hence, if we need to extend relational triple extraction into a new domain, a few examples are needed to activate the system in the new domain without retraining the model. By formulating this FSL relational triple extraction, we can significantly reduce the annotation cost and training cost while maintaining highly accurate results.

Though methods of few-shot learning develop fast in recent yeas, most of these works concentrate on single tasks such as relation extraction and text classification \cite{Geng2019InductionNF,Ye2019MultiLevelMA}. However, the effect of joint extraction of entities and relations on low-resource scenarios is still not well understood, which are two subtasks belonging to relational triple extraction.  Unlike extraction for each single task, joint entity and relation extraction is more challenging, as entity and relations have implicit  correlations, which cannot be ignored.

To address this issue, we propose a Multi-Prototype Embedding  network (MPE) model to extract the few-shot relational triples,   inspired by the prototypical network \cite{snell2017prototypical}.  To be specific, we utilize two kinds of prototypes regarding both entities and relations. Note that  entity pairs and relations have explicit knowledge constraints \cite{bordes2013translating}, such as the \emph{Born\_in} relation suggests that the type of \emph{head entity} must be \emph{PERSON}, and vice versa. Based on those observations and motivated by the knowledge graph embedding \cite{xie2016representation}, we introduce the hybrid prototypical learning to explicitly inject knowledge constraints.  We firstly learn entity and relation prototypes and then leverage translation constraint in hyperspace to regularize prototype embedding. Note that such knowledge-aware regularization not only injects prior knowledge from the external knowledge graph, but also leads to a more smooth and representative prototype for few-shot extraction. Moreover, we introduce prototype-regularization considering both intramural and mutual similarities between different prototypes.  Experimental results on the FewRel dataset \cite{han2018fewrel} demonstrate that our approach outperforms baseline models in the few-shot setting.

To summarize, our main contributions include:
\begin{itemize}
\item We study the few-shot relational triple extraction problem and provide a baseline for this new research direction. To our best knowledge, this is a new branch of research that has not been explored.

\item We propose a novel Multi-Prototype Embedding approach with hybrid prototype learning and prototype-aware regularization, which bridge text and knowledge for few-shot relational extraction.

\item Extensive experimental results on the FewRel dataset demonstrate the effectiveness of our method.

\end{itemize}

\section{Related Work}

Two main directions have been proposed for relational triple extraction, which has two subtasks: entity extraction and relation extraction, namely, pipeline \cite{lin2016neural,Trisedya2019NeuralRE,wang2020finding,zhang2020can,nan2020reasoning} and joint learning methods \cite{Bekoulis2018JointER,Nayak2020EffectiveMO,ye2020contrastive}.
The pipeline model can be more flexible because it extracts entity pairs and relations sequentially, but this design will lead to error propagation \cite{zhang2018attention}. Meanwhile, joint relational triple extraction models can solve this problem well by extracting triples end-to-end, and the interaction between entities and relations can be realized within the model, which makes the performance of the two mutually enhanced.

However, due to the ``data-hungry" attribute of conventional neural networks, these relational triple extraction models need a large amount of data for training. Thus, lots of efforts \cite{zhang2019long,yu2020devil,zhang2020relation} have been devoted to few-shot learning, \cite{han2018fewrel} presents a few-shot relation extraction datasets to promote the research of information extraction in few-shot scenarios and adapt some few-shot learning methods \cite{munkhdalai2017meta,satorras2018few,mishra2017simple,deng2020meta} for this task. Among these models, the prototypical network \cite{snell2017prototypical} achieves comparable results on several few-shot learning benchmarks; meanwhile, it is simple and effective. This model assumes that each class exists a prototype, and it tries to find the prototypes for classes from supporting instances and compares the distance between the query instance under a particular distance metric. In natural language processing, \cite{gao2019hybrid} first proposes a hybrid attention-based prototypical network for few-shot relation extraction.  \cite{fritzler2019few} proposes to utilize the prototypical network to tackle the few-shot named entity recognition.   \cite{hou2020few} proposes a collapsed dependency transfer mechanism and a Label-enhanced Task-Adaptive Projection Network (L-TapNet) for few-shot slot filing.  However, all previous few-shot works mainly consider single tasks, while relational triple extraction should take both entity and relation into consideration.  To the best of our knowledge, we are the first approach for the few-shot relational triple extraction, which addresses both entities and relations. 

Our work is motivated by knowledge graph embedding \cite{xie2016representation} such as TransE \cite{bordes2013translating} from Knowledge graph (KG), which  is composed of many relational triples like {\it $\langle$head, relation, tail$\rangle$}. TransE is first proposed by \cite{bordes2013translating} to encode triples into a continuous low-dimensional space, which is based on the translation law $h + r \approx t$. Many follow-up works like TransH \cite{wang2014knowledge}, DistMult ~\cite{yang2014embedding}, and TransR ~\cite{lin2015learning}, propose advanced methods of translation by introducing different embedding spaces. In few-shot settings, it is extremely challenging to inject implicit knowledge constrains in vector space. Such simple yet effective knowledge constraints provide an intuitive solution.

\section{Methodologies}

\subsection{Problem Definition}

In few-shot relational triple extraction task, we are given two datesets, $\mathcal{D}_{meta-train}$ and $\mathcal{D}_{meta-test}$. Each dataset consists of a set of samples ({\it x, t}), where {\it x} is a sentence composed of {\it N} words, and {\it t} indicates  relational triple extracted from {\it x}. The form of {\it t}  is {\it $\langle$head, relation, tail$\rangle$}, where {\it head} and {\it tail} are entity pairs associated with the {\it relation}. These two datasets have their own relation domain spaces that are disjoint with each other.
In few-shot settings, $\mathcal{D}_{meta-test}$ is split into two parts: $\mathcal{D}_{test-support}$ and $\mathcal{D}_{test-query}$. Due to entity pair types can be determined by the relation categories, e.g. the \emph{Born\_in} relation suggests that the type of \emph{head} might be \emph{PERSON} and {\it tail} might be {\it LOCATION}, we are able to determine the classification of triples only by specifying the categories of the relations. Therefore if $\mathcal{D}_{test-support}$ contains {\it K} labeled samples for each of {\it N} relation classes, this target few-shot problem is named {\it N}-way-{\it K}-shot. $\mathcal{D}_{test-query}$ contains test samples, each should be labeled with one of  {\it N} relation classes, and associated entity pairs also need to be extracted correctly.

It is non-trivial to train a good model from scratch using $\mathcal{D}_{test-support}$ and evaluate its performance on $\mathcal{D}_{test-query}$, limited by the number of test-support samples (i.e.,., {\it N $\times$ K}). Inspired by an important machine learning principle that test and train conditions must match, we can also split $\mathcal{D}_{meta-train}$ into two parts, $\mathcal{D}_{train-support}$ and $\mathcal{D}_{train-query}$, and mimic the few-shot settings at the training stage. In each training iteration, {\it N} triple categories are randomly selected from $\mathcal{D}_{train-support}$, and {\it K} support instances are randomly selected from each of {\it N} triple categories. In this way, we construct the train-support set $S=\{s^i_k; i=1,\dots,N, k=1,\dots,K\}$, where $s^i_k$ is the {\it k}-th instance in triple category {\it i}. Meanwhile, we randomly select {\it R} samples from the remaining samples of those {\it N} triple categories and construct the train-query set $Q=\{(q_j,t_j); j=1,\dots,R\}$, where $t_j$ is the triple extracted from instance $q_j$. Our goal is to optimize the following function:

\begin{equation}
  L = - \frac1{R} \sum_{(q,t) \in Q} P(t|S,q)
\end{equation}

Where $P(t|S,q)$ is the probability of gold standard relational triples.

\subsection{Framework Overview}

\begin{figure*}
  \centering 
  \includegraphics[width=0.9\textwidth]{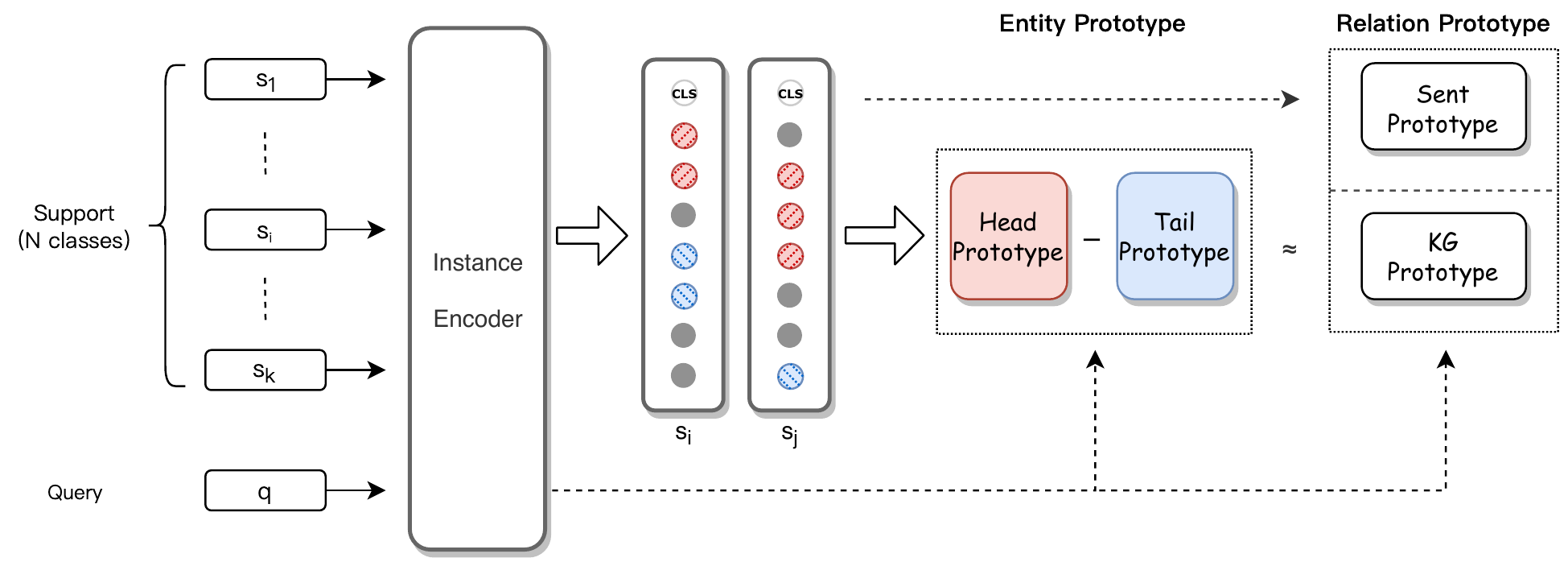} 
  \caption{Overview of our proposed Multi-Prototype Embedding (MPE) model. Best view in color.}
  \label{figure:model}
\end{figure*}

In this section, we will introduce our proposed Multi-Prototype Embedding (MPE)  model for few-shot relational triple extraction. For brevity, we will temporarily study a sentence with one relation and associated entity pairs. 
The framework of our proposed model is shown in Fig.~\ref{figure:model}, which has three main modules.

\begin{itemize}
\item \textbf{Instance Encoder.} We utilize the pre-trained language model BERT \cite{devlin2018bert} to encode sentence, which adopts multi-head attention to learn contextual representations. Note that any other encoders such Roberta \cite{liu2019roberta} and XLNet \cite{yang2019xlnet} can also be applied.

\item \textbf{Hybrid Prototype Learning.} After obtaining entity pairs representations of each sentence used by sequence labeling methods, we can get entity prototypes in support set, and then construct relation prototype based on knowledge graph constraint, which takes the interaction between entity pairs and relations into account.

\item \textbf{Prototype-Aware Regularization.} To further enhance the prototype learning, we optimize the position of prototypes in representation spaces. We make the distance between each prototype and related instances closer and distract those prototypes with different types. 
\end{itemize}

\subsection{Instance Encoder}

For each sentence $x = \{w_1, w_2, \dots, w_n\}$ in the  support or query dataset, where $w_i \in x$ is the word token in sentence $x$, we first construct input sentence in the form: \{[CLS], $w_1, w_2,$ $\dots, w_n,$ [SEP]\}, in order to match the input of BERT ~\cite{devlin2018bert}. The pre-trained language model has been shown to be effective in many NLP tasks. [CLS] token is used to represent the entire sentence information, and [SEP] is the end token of sentence. 
After multi-head attention ~\cite{vaswani2017attention} calculation, we can get sentence contextual embeddings $\mathcal{B} = \{h_0,$ $h_1,$ $h_2,$ $\dots,$ $h_n,$ $h_{n+1}\}$, where $ \mathcal{B} \in \mathbb{R}^{d_{n+2} \times d_b} $, $d_b$ is BERT pre-defined hidden size, $h_0$ is [CLS] token embedding, $h_{n+1}$ is [SEP] token embedding, and $h_i, i \in [1, n]$ is each token embedding in sentence. Note that $n$ can be different from input sentence length because of tokenizer (e.g., byte-pair-encoding) might split words into sub-tokens.

\subsection{Hybrid Prototypical Learning}

\textbf{Entity Prototype Learning.} During training stages, sentence representations in support datasets are first used to construct the entity pairs prototypes. We build an entity labeling set S = \{B-Head, I-Head, B-Tail, I-Tail, O, X\} to label out each token in the sentence, where B-Head, I-Head indicate head entity positions, B-Tail, I-Tail indicate tail entity positions, O means other tagging labels, and X is any remaining fragments of tokens split by the tokenizer. 

We utilize Conditional Random Field (CRF) \cite{lafferty2001conditional} for sequence labeling as it models the constraints between labels, which is more convenient in few-shot learning scenarios.  Let $y=$ $\{y_0,$ $y_1,$ $y_2,$ $\dots,$ $y_n,$ $y_{n+1}\}$, where $y_0$ is [CLS] token label which means the start of sentence, $y_{n+1}$ is [SEP] token label which means the end of sentence, and $y_i, i \in [1, n]$ is each token label of sentence in entity labelling set.
CRF uses emission and transition scores to combine local and global information, in our model, score of this sequence is evaluated as:

\begin{align}
  Score(x,y) = \sum^{N+1}_{i=0} E_{y_i,i} + \sum^N_{j=0} T_{y_i,y_{j+1}}
\end{align}

Let $\mathcal{Y}_\mathcal{X}$ indicate the exponential space of all possible labelings of this sequence $x$. The probability of a specific labeling $y \in \mathcal{Y}_\mathcal{X}$ is  evaluated as:


\begin{align}
  p(y|x) = \frac{e^{Score(x,y)}}{\sum_{y \in \mathcal{Y}_\mathcal{X}} e^{Score(x,y)}}
\end{align}


We name the CRF-based sequence labeling loss as $loss_{crf}$ and minimize it during training stage.

After the above instance encoder and sequence labeling, we can obtain the head and tail representation to match the entities between the query and support set. Due to the variable length of entity words, we only use the first token representation of each entity word as head/tail embeddings, which is also used in  \cite{soares2019matching}.
For measuring the distance between samples in query set and support set, we need compute a representative vector, called prototype, for each class $t \in T$ in the support set $S$ from its instances' vectors. The original Prototypical Network \cite{snell2017prototypical} hypothesis that all instance vectors are equally important, so it aggregates all the representation vectors of the instance of class $t_i$, and then perform averaging over all vectors as follows:

\begin{align}
    head_{proto} = \frac{1}{|S_k|} \sum_{head_i \in S_k} head_i  \qquad
    tail_{proto} = \frac{1}{|S_k|} \sum_{tail_i \in S_k} tail_i 
\end{align}

where $head_i, tail_i$ are each sentence's entity pairs representations. Intuitively, the instances of a given relation may be quite different. Thus, we propose to adopt weighted sum prototype, named Proto+Att network inspired by \cite{gao2019hybrid}. The weights are obtained by attention mechanism according to the representational vector of the query $Q$ as follow:

\begin{align}
    head_{proto} = \frac{1}{|S_k|} \sum_{head_i \in S_k} \alpha_h head_i  \qquad
    tail_{proto} = \frac{1}{|S_k|} \sum_{tail_i \in S_k} \alpha_t tail_i 
\end{align}

where

\begin{align}
  \begin{split}
    \alpha_h &= \frac{\text{exp}(e_{h_i})}{\sum_{m=1}^k \text{exp}(e_{h_m})}  \;\quad  e_{h_i} = head_{proto}^T Q
    \\
    \alpha_t &= \frac{exp(e_{t_j})}{\sum_{n=1}^k exp(e_{t_n})}   \qquad  e_{t_j} = tail_{proto}^T Q
  \end{split}
\end{align}

Specifically, we use Euclidean distance $d(\mathbf{z}-\mathbf{z}') = \|\mathbf{z}-\mathbf{z}'\|^2$, to calculate the distance between  entity prototypes and instances in query set, and minimize this distance as $loss_{entity}$.

\paragraph{Relation Prototype Learning.} This module computes relation prototypes associated with each entity pair.  On the one hand,  the first token [CLS] in the sentence representation can represent the whole sentence information. So like the above entity prototypes calculation, we can get sentence prototypes $sent_{proto}$, used by this sentence information in support set.

On the other hand, knowledge graph representation learning inspires us to learn a translation law $h + r \approx t$ \cite{bordes2013translating} on a continuous low-dimensional space, where $h,r,t$ describe the head entity, the relation and the tail entity respectively. So we use $head_{proto}$ and $tail_{proto}$ to construct knowledge graph prototype $kg_{proto}$, which takes the interaction between entities and relations into consideration as follows:
 
\begin{align}
  kg_{proto} = |head_{proto} - tail_{proto}| W_r
\end{align}

Finally, we combine the prototype of sentence represent ions $sent{proto}$ and prototype from knowledge constrains between entity pairs $kg_{proto}$ to form the relation prototype as follows:

\begin{equation}
    relation_{proto} = [sent_{proto}; kg_{proto}],
    \label{eq_proto}
\end{equation}

Where $[;]$ refers to the feature vector concatenation. Similar to the entity prototype, we use Euclidean distance to calculate the distance between relation prototype $relation_{proto}$ and the sentence in the query set $Q$, and minimize this distance as $loss_{relation}$.

\subsection{Prototype-Aware Regularization}
Previous few-shot learning approaches \cite{Ye2019MultiLevelMA} have shown that if the representations of all support instances in a class are far away from each other, it could become difficult for the derived class prototype to capture the common characteristics of all support instances. Therefore, we propose prototype-aware regularization to optimize prototype learning.  Intuitively, we argue that the representational vectors (e.g, sentence representations/prototypes) of the same class should be close to each other; the prototypes of different types should be located far from each other in the prototypical space. Specifically, We use Euclidean and Cosine distance to measure these similarities, and optimize the prototype represetations as follows:

\begin{align}
    loss_{intra} = \frac{1}{NK} \sum_{i=1}^N \sum_{k=1}^K \|x_i^k-p_i^k\|_2^2 \qquad
    loss_{inter} = 1 - \frac{1}{N} \sum_{i=1}^N \sum_{j={i+1}}^N cosine(p_i, p_j)
\end{align}

where $x_i$ is each sentence representation, $p_i$ is associated prototypes, $loss_{intra}$ and $loss_{inter}$ are two different prototype-aware regularization functions. The overall regularizationn loss is: $loss_{regular}$ = $loss_{intra}$ +  $\alpha loss_{inter}$, and $\alpha$ is hyperparameter.

The overall objective of the optimization is as follows:

\begin{align}
  L =  loss_{crf} + \beta loss_{entity} + \gamma loss_{relation} + \delta loss_{regular}
\end{align}

where $\beta$, $\gamma$ and $\delta$ are the trade-off parameters.

\section{Experiments}

\subsection{Datasets}
We conduct experiments on the public dataset FewRel\footnote{https://www.zhuhao.me/fewrel/} ~\cite{han2018fewrel}, which is derived from Wikipedia and annotated by crowd workers.
FewRel releases 80 relation categories, and each relation has 700 samples.
We reconstruct the FewRel dataset to satisfy the few-shot relational triple extraction task.
Our input information has only one sentence, and the required output is the relation and related entity pairs, which is a complete knowledge triple in the scheme of {\it $\langle$head, relation, tail$\rangle$}.
In our experiments, we randomly select 50 relations for training, 15 for validation, and the rest 15 relation types for testing. Note that there are no overlapping types between these three datasets.

\subsection{Settings}

\begin{wraptable}{r}{0.5\textwidth}
\centering 
\resizebox{0.48\textwidth}{!}{
\begin{tabular}{ccc}
\toprule
Component  &  Parameter  &  Value \\
\midrule
\multirow{2}{*}{BERT} & type        &  base-uncased \\
                      & hidden size & 768           \\

\midrule
\multirow{2}{*}{Dataset} & $N_{train}$ & 20\\
                         & $R_{query}$ &  5\\
\midrule
\multirow{4}{*}{learning rate} & init (proto) & 0.1    \\
                               & init (BERT)  & 0.0005 \\
                               & weight decay & 1/3    \\
                               & decay steps  & 2000   \\
\midrule
\multirow{4}{*}{loss} & $\alpha$  & 0.75 \\
                      & $\beta$   & 0.5  \\
                      & $\gamma$  & 0.8  \\
                      & $\delta$  & 1    \\
\bottomrule
\end{tabular}
}
\caption{Hyper-parameters of our approach.}
\label{hyperparameters}
\end{wraptable}

We implement our approach with Pytorch \cite{paszke2019pytorch}. We employed mini-batch stochastic gradient descent (SGD) \cite{Bottou2010LargeScaleML} with the initial learning rate of $1e^{-1}$.
The learning rate was decayed to one third with every 2000 steps, and we train 30,000 iterations.
The dropout rate of 0.2 is used to avoid over-fitting.
Previous study ~\cite{snell2017prototypical} found that models
trained on more laborious tasks may achieve better performances than using the same configurations at
both training and test stages. Therefore, we set $N = 20$ to construct the train-support sets
for 5-way and 10-way tasks. Furthermore, in each step, we sample 5 instances for query datasets. We utilize grid-search on valid set to tune hyperparameters.  All of the hyperparameters used in our experiments are listed in Table \ref{hyperparameters}.

We consider two types of few-shot relational triple extraction tasks in our experiments: 5-way 5-shot and 10-way 10-shot. We evaluate the performance of the entity, relation, and triple with the micro F1 score. To be specific, the entity performance refers to that the entity's span and span type are correctly predicted, the relation performance means that the relation of the entity pairs is correctly classified. Moreover, the triple performance means that the entity pair and associated relation are all matched correctly.

\subsection{Baselines}
We compared our model with baselines of supervised approaches and few-shot learning methods:

Supervised Learning.  We utilize BERT \cite{devlin2018bert} with fine-tuning (Finetune) as supervised learning baselines. We finetune BERT with a batch size of 16 for 100 iterations.  

Few-shot Leaning. We apply Matching Network (MatchingNet) \cite{Vinyals2016MatchingNF}, Relation Network (RelationNet) \cite{sung2018learning},  vanilla Prototypical Network (Proto) \cite{snell2017prototypical}  and Prototypical Network with attention (Proto+Att) \cite{Ye2019MultiLevelMA} as few-shot baselines. We only utilize the sentence prototypes  $sen_{proto}$ in few-shot baselines which do not take the implicit knowledge into consideration.

\subsection{Overall Evaluation  Results}

\begin{table*}
\centering

\resizebox{\textwidth}{!}{
\begin{tabular}{l|ccc|ccc}  
\toprule
\multirow{2}{*}{\textbf{Model}}  & \multicolumn{3}{c|}{5-Way-5-Shot} & \multicolumn{3}{c}{10-Way-10-Shot}\\

\cmidrule{2-7}   & Entity & Relation & Triple   & Entity & Relation & Triple\\
\midrule 

Finetune    &  6.57$\pm$0.52 & 71.36$\pm$0.49 & 4.71$\pm$0.96
            &  4.36$\pm$0.63 & 63.83$\pm$0.60 & 2.94$\pm$0.77\\ 
\midrule 
MatchNet    & 12.30$\pm$0.74 & 84.15$\pm$0.28 & 10.13$\pm$0.43
            &  5.94$\pm$1.23 & 79.34$\pm$0.51 & 4.40$\pm$1.02 \\
RelationNet & 11.87$\pm$1.61 & 88.73$\pm$0.11 & 9.91$\pm$0.28
            &  8.24$\pm$0.79 & 82.40$\pm$0.37 & 6.65$\pm$0.33 \\
Proto       & 15.43$\pm$0.50 & 87.10$\pm$0.25 & 14.18$\pm$0.25
            &  7.76$\pm$0.51 & 80.46$\pm$0.33 & 6.53$\pm$0.60 \\
Proto+Att   & 19.84$\pm$0.83 & 89.29$\pm$0.36 & 18.20$\pm$0.46
            & 11.66$\pm$0.39 & 82.95$\pm$0.19 & 10.55$\pm$0.31 \\
\midrule
MPE         & \textbf{25.03$\pm$1.24} & \textbf{93.81$\pm$0.31} & \textbf{23.34$\pm$0.79}
            & \textbf{14.85$\pm$0.80} & \textbf{84.58$\pm$0.32} & \textbf{12.08$\pm$0.83} \\

\bottomrule
\end{tabular}
}
\caption{F1 score on the FewRel test set.
}
\label{result}
\end{table*}

The first line of Table \ref{result} shows the performance of our model on the FewRel test set.
From the results, we observe that:

1) Our approach MPE achieve the best performance in few-shot setting compared with all baselines (about absolute 5\% improvement than Proto+Att in 5-way-5-shot), which demonstrates that the multi-prototype leveraging both text and knowledge is effective.

2) Entity recognition performs much worse than relation extraction in few-shot settings, as sequence labeling is more challenging than classification tasks, and the empirical results also observed by  \cite{hou2020few}. More studies need to be taken to handle the challenging few-shot entity recognition task. 

3) Proto+Att achieve better performance than Proto, which reveals that different instances have different contribution to prototype learning. 

4) The overall performance is far from satisfactory, which need more future works to be taken into consideration.

\subsection{Ablation Study}

\begin{wraptable}{r}{0.5\textwidth}
\centering
\begin{tabular}{l|ccc}
\toprule
\multirow{2}{*}{Muitl-Proto}  & \multicolumn{3}{c}{10-Way-10-Shot} \\
\cmidrule{2-4}  & Entity & Relation & Triple  \\
\midrule
MPE       & 14.85 & 84.58 & 12.08 \\
\midrule 
w/o CRF   &  8.03 & 83.55 &  6.96 \\
w/o Att   & 12.27 & 82.42 & 10.32 \\
w/o intra & 11.33 & 80.49 &  9.35 \\
w/o inter & 12.86 & 81.22 & 10.50 \\
\bottomrule
\end{tabular}
\caption{Ablation study.}
\label{ablation}
\end{wraptable}

%



We further analyze the different modules of our approach by taking ablation studies, as shown in Table \ref{ablation}. w/o CRF implied without the CRF decoder; w/o Att implied without the attention in prototypical learning; w/o intra implied without the intra-  constrains  between instances and prototypes ;  w/o inter implied without the inter-  constrains  between prototypes. From Table \ref{ablation}, we observe that:

1) All approaches without different modules  obtain performance decays, and w/o CRF has significant performance decay than w/o Att, w/o intra, and w/o inter, which demonstrates that the efficacy of CRF is more critical in few-shot relational triple extraction. 

2) w/o intra or w/o inter has more performance drop compared with w/o Att, which also illustrates that prototype-aware regularization benefits the prototype learning.  

\begin{figure*}[h]
\centering

\subfigure[5-Way-5-Shot] { 
  \includegraphics[scale=0.34]{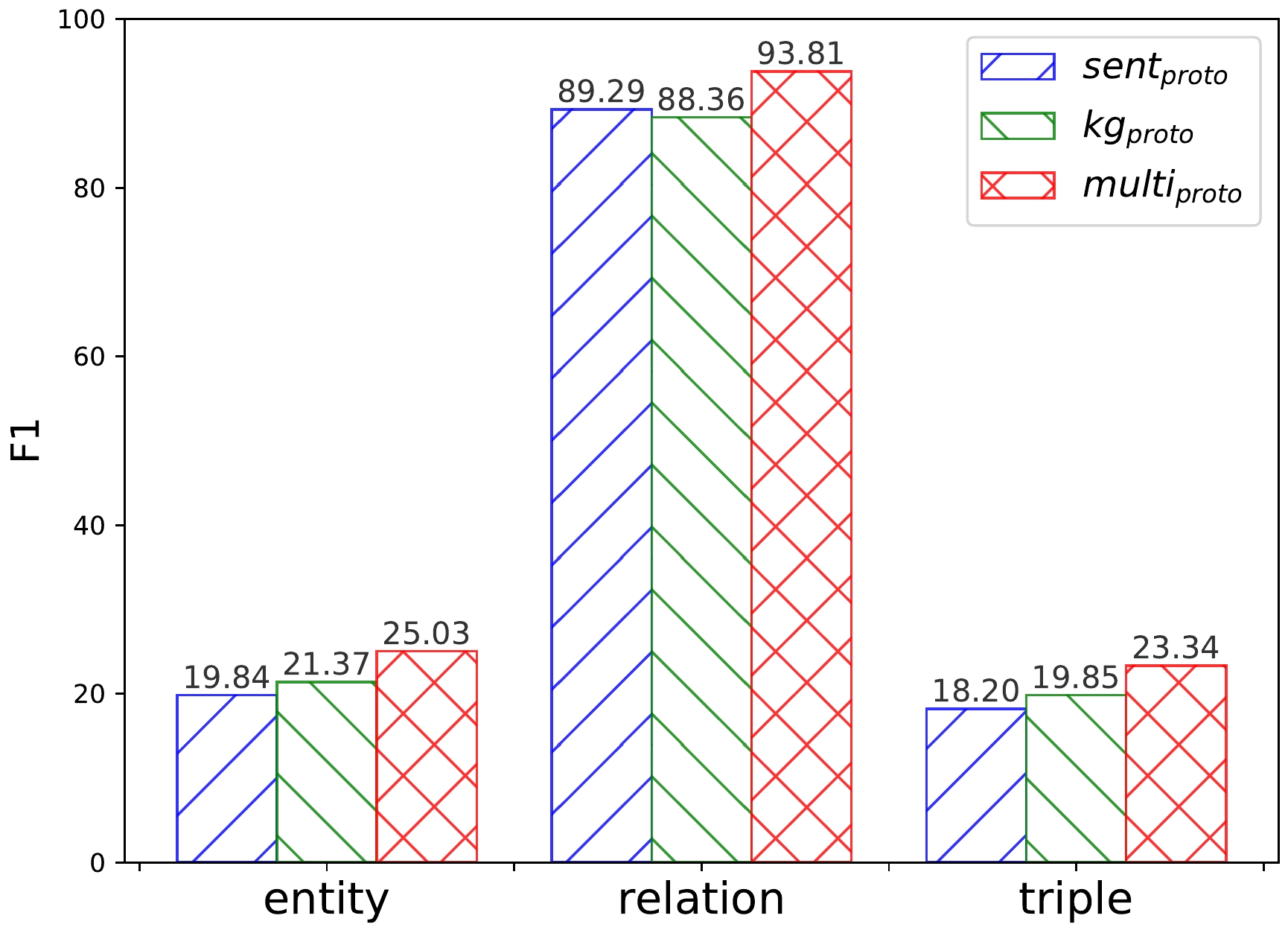}  
}
\subfigure[10-Way-10-Shot] { 
\includegraphics[scale=0.34]{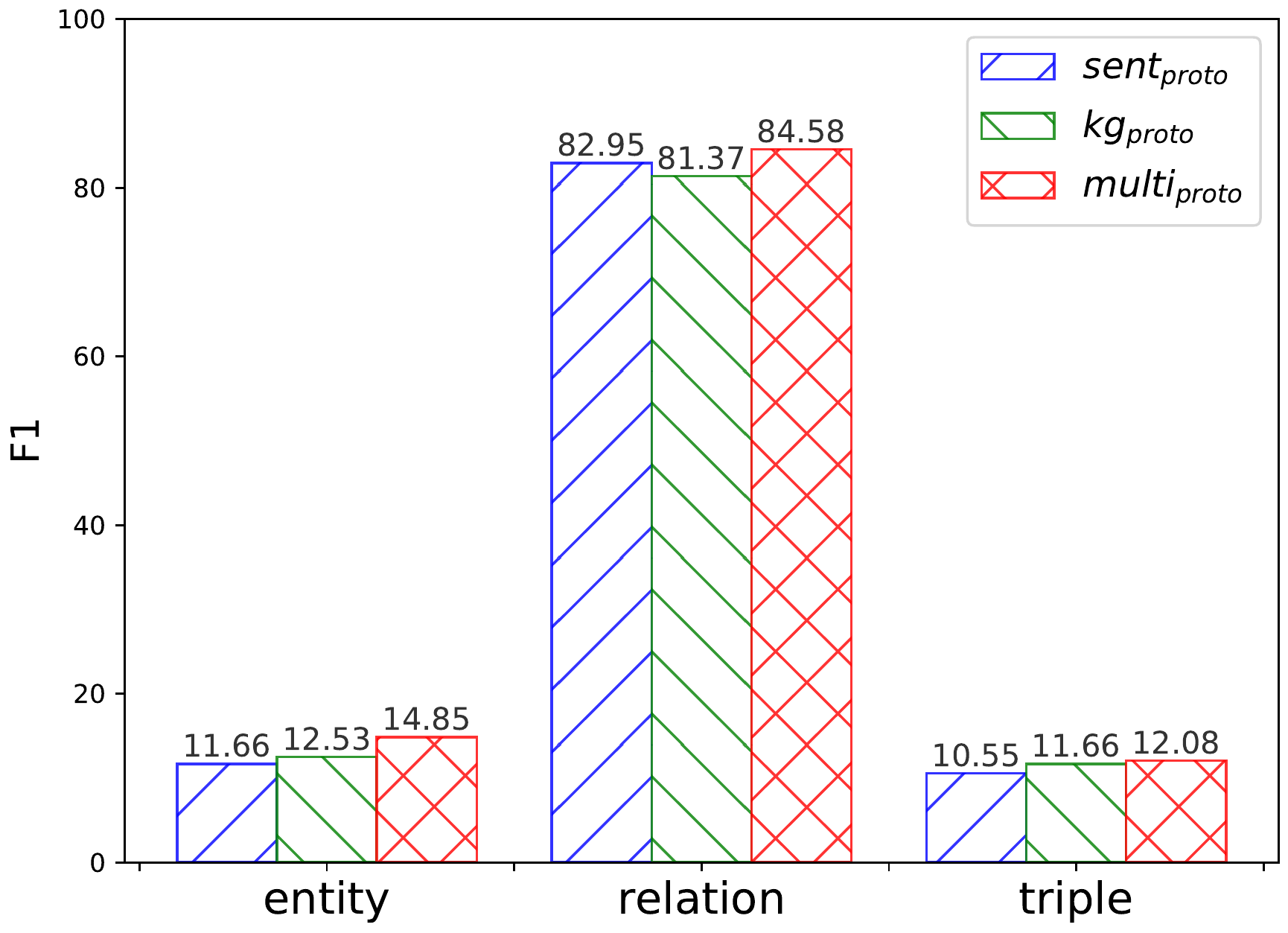}
}
\caption{Evaluation results of models with $sent_{proto}$, $kg_{proto}$ and $multi_{proto}$.}
\label{bar}
\end{figure*}

From Figure \ref{bar}, we observe that the $multi_{proto}$ achieves better performance than $sen_{proto}$ and $kg_{proto}$, and $kg_{proto}$ is more advantageous than $sen_{proto}$ for entity extraction, which further indicates that such knowledge constrains is beneficial.

In summary, we observe that entity recognition is more difficult than relation extraction in few-shot settings and the implicit correlation between them contribute to the performance.

\begin{table*}
  \centering 
  \resizebox{\textwidth}{!}{
\begin{tabular}{l}
  \toprule
  {\bf Query Instance}\\
  \midrule
  \underline{instance \#1} \ \quad
  Delias mandaya is a species of pierine butterfly endemic to Mindanao, in the Philippines.\\

  extracted triplet: 
  {\it $\langle${\color{reds}{Mindanao}}, contains administrative territorial entities, {\color{blues}{Philippines}}$\rangle$} \\

  ground truth: \ \quad
  {\it $\langle${\color{reds}{Mindanao}}, country, 
  {\color{blues}{Philippines}}$\rangle$} \vspace{1 ex} \\
   
  \underline{instance \#2} \;
  Hamilton Hyde Kellogg was the Fifth Bishop of Minnesota in The Episcopal Church.\\

  extracted triplet: 
  {\it $\langle${\color{reds}{Hamilton Hyde}}, religion, 
  {\color{blues}{Church}}$\rangle$} \\

  ground truth: \ \quad
  {\it $\langle${\color{reds}{Hamilton Hyde Kellogg}}, religion,}  {\it {\color{blues}{The Episcopal Church}}$\rangle$} 
  \vspace{1 ex} \\

  \underline{instance \#3} \;
  His family has roots in the earliest Catholic presence in the United States west of the\\ Appalachian Mountains; among his relatives are Martin John Spalding and John Lancaster Spalding. \\

  extracted triplet: 
  {\it $\langle${\color{reds}{John Lancaster Spalding}}, religion, {\color{blues}{Catholic}}$\rangle$} \\

  ground truth: \ \quad
  {\it $\langle${\color{reds}{Martin John Spalding}}, religion, 
  {\color{blues}{Catholic}}$\rangle$}  \\

  \bottomrule
  \end{tabular}
  }
\caption{Error analysis.}
\label{errorAnylysis}
\end{table*}

\subsection{Error Analysis}

To further analyze the drawbacks of our approach and promote future works of few-shot relational extraction, we random select instances and conduct error analysis, 
as shown in Table \ref{errorAnylysis}:

Distract Context. As instance \#1 shows, we observe that our approach may fail to those ambiguous contexts that may be expressed in a similar context but differ only in the fine-grained type of entities. We argue that this may be caused by the unbalanced learning problems that models tend to classify the sentence with similar context to high-frequency relations. 

Wrong Boundaries. As instance \#2 shows, we observe that lots of extracted triples have incorrect boundaries, which further demonstrates the difficulty of entity recognition in the few-shot setting. More future works should be focused on the direction of few-shot sequence labeling. 

Wrong Triples. As instance \#3 shows, we observe that lots of extracted triples have entities that do not exist in the gold standard set. Generally, this is mostly happening in the sentence with multiple triples. Note the FewRel dataset does not have those labeled triples, and part of those cases is correct.  

\section{Conclusion and Future Work}

In this paper, we study the few-shot relational triple extraction problem and propose a novel multi-prototype embedding network that bridge text representation learning and knowledge constraints. 
Extensive experimental results prove that our model is effective, but remains challenges. 
Those empirical findings shed light on promising future directions, including 1) enhancing entity recognition with effective sequence decoders;  2) studying few-shot relational triple extraction with more triples in a single sentence; 3) injecting logic rules to enable robust extraction; and 4) developing few-shot relational triple extraction benchmarks. 

\section*{Acknowledgments}
We  want to express gratitude to the anonymous reviewers for their hard work and kind comments, which will further improve our work in the future. This work is funded by NSFCU19B2027/91846204/61473260, national key research program SQ2018YFC000004/2018YFB1402800, Alibaba CangJingGe (Knowledge Engine) Research Plan.

\bibliographystyle{coling}
\bibliography{myref}

\end{document}